\documentclass[journal]{IEEEtran}
\usepackage[caption=false,font=normalsize,labelfont=sf,textfont=sf]{subfig}
\usepackage[pdftex]{graphicx}
\usepackage{stfloats}
\usepackage{rotating}
\usepackage{graphicx}  
\usepackage{amsmath}
\usepackage{booktabs}
\usepackage{amssymb}
\usepackage{textcomp,booktabs}
\usepackage{multirow}
\usepackage[usenames]{color}
\usepackage{wrapfig}
\usepackage{longtable}
\usepackage{colortbl,booktabs}
\usepackage{verbatim}

\begin{document}

\title{Rethinking of the Image Salient Object Detection:
 Object-level Semantic Saliency Re-ranking First, Pixel-wise Saliency Refinement Latter}

\author{Guangxiao Ma$^{1}$
        ~~~~Shuai Li$^{1}$
        ~~~~Chenglizhao Chen$^{1,2*}$\thanks{Corresponding author: Chenglizhao Chen, cclz123@163.com. Guangxiao Ma and Shuai Li contributed equally to this work.}
        ~~~~Aimin Hao$^{1}$
        ~~~~Hong Qin$^3$
        \\ $^1$Qingdao Research Institute \& State Key Laboratory of VRTS, Beihang University\\
        $^2$Qingdao University~~~~~$^3$Stony Brook University\\
        Source Code and Data, https://github.com/gxma/TIP\_RISOD}

\markboth{IEEE Transactions on Image Processing,~Vol.~x, No.~x, x~x}%
{Shell \MakeLowercase{\textit{et al.}}: Rethinking of the Image Salient Object Detection: Object-level Semantic Saliency Re-ranking First, Pixel-wise Saliency Refinement Latter}

\maketitle

\begin{abstract}
The real human attention is an interactive activity between our visual system and our brain, using both low-level visual stimulus and high-level semantic information.
Previous image salient object detection (SOD) works conduct their saliency predictions in a multi-task manner, i.e., performing pixel-wise saliency regression and segmentation-like saliency refinement at the same time, which degenerates their feature backbones in revealing semantic information.
However, given an image, we tend to pay more attention to those regions which are semantically salient even in the case that these regions are perceptually not the most salient ones at first glance.
In this paper, we divide the SOD problem into two sequential tasks: 1) we propose a lightweight, weakly supervised deep network to coarsely locate those semantically salient regions first; 2) then, as a post-processing procedure, we selectively fuse multiple off-the-shelf deep models on these semantically salient regions as the pixel-wise saliency refinement.
In sharp contrast to the state-of-the-art (SOTA) methods that focus on learning pixel-wise saliency in ``single image'' using perceptual clues mainly, our method has investigated the ``object-level semantic ranks between multiple images'', of which the methodology is more consistent with the real human attention mechanism.
Our method is simple yet effective, which is the first attempt to consider the salient object detection mainly as an object-level semantic re-ranking problem.
\end{abstract}



\section{Introduction}
Salient object detection aims to fast locate the most eye-catching objects in a given scene.
As an early perceptual tool to support the high-level recognition activities of human brain, a desired salient object detection (SOD) method should be lightweight designed~\cite{fan2019SOC,fang2019visual,chen2019improved,CC2019CVPR}.
After entering the deep learning era, the state-of-the-art (SOTA) SOD methods have achieved high quality detection performance in an extremely fast end-to-end manner.

In general, the SOTA network architectures~\cite{zhuge2019deep,zeng2019multi,CC2020TIP,li2019motion} frequently consist of an off-the-shelf feature backbone (e.g., the vanilla VGG16 or ResNet50), followed by an saliency estimation sub-net with elaborate network design.
The backbone networks are pre-trained via large scale training set that is abundant in semantic information (e.g., object categories), aiming to span a high discriminative feature space for its subsequent saliency estimation sub-net.
Nevertheless, the saliency estimation sub-nets in the SOTA methods were applied to a multi-task job, i.e., performing both the pixel-wise saliency regression and the segmentation-like saliency refinement simultaneously, failing to make full use of the semantic information provided by its precedent feature backbone.

\begin{figure}[!t]
\centering
\includegraphics[width=1\columnwidth]{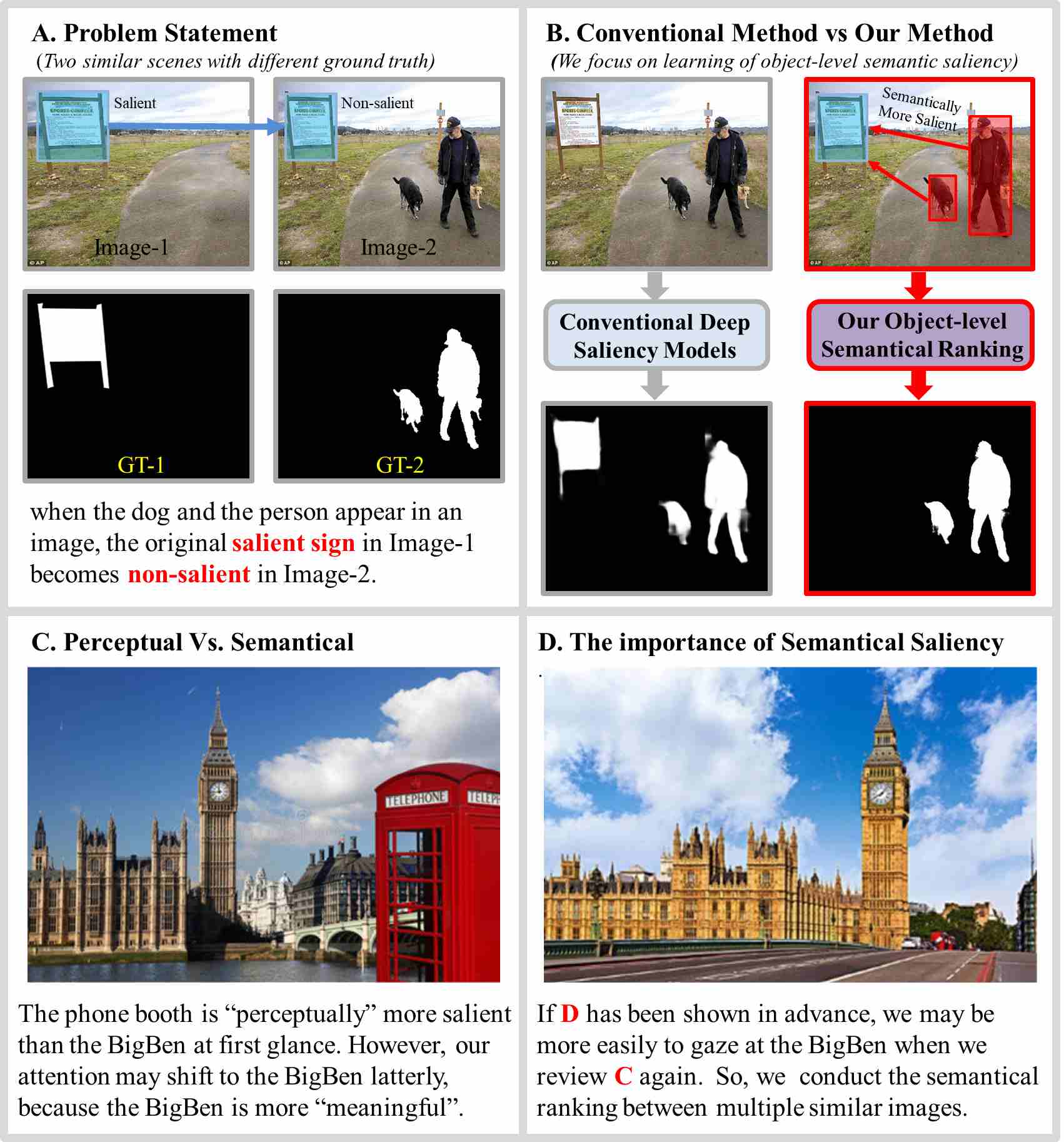}
\caption{The conventional methods consider the image saliency mainly from the perceptual perspective, while the real human vision system may pay more attention to those semantically salient objects.
Specifically, as shown in C and D, simply following the conventional definition to regard the phone booth as salient object seems questionable. This is because, from the perspective the SOD related applications such as Image Compression, it may be more useful to detect the meaningful BigBen as the salient object. We aim to emphasis the importance of the semantic saliency, which may be more useful than the perceptual saliency to predict where will we really look at. Thus, this paper provides a novel weakly supervised scheme to learn the object-level semantic saliency, which is more consistent with the real human visual system in \textbf{salient object localization} (averagely \underline{8\%} improvement in maxFm, Table.~\ref{table:DifferentAq}), producing much better \textbf{saliency maps} ({averagely \underline{4\%} improvement in maxFm}, Table.~\ref{table:ComparisonDifferentModels}).}
\label{fig:Motivation}
\end{figure}

In fact, the real human attention mechanism~\cite{CC2019TMM2} is not solely dependent on the low-level visual stimulus provided by the human visual system.
On the contrary, the high-level semantic information rooted in our brain plays an extremely important role to determine where we look at.
For example, as shown in Fig.~\ref{fig:Motivation}-A, the salient \emph{road sign} in the image-1 will become non-salient when a person with a dog appears in the image-2.
In such case, though the perceptual aspects of the \emph{road sign}, e.g., appearance, location and contrast, are not even changed, we may still tend to pay more attention to those semantically more salient regions, i.e., the person and the dog in image-2); however, the conventional methods~\cite{hu2018recurrently,HouPami19Dss} may still assign large saliency value to the \emph{road sign} incorrectly (see left column in Fig.~\ref{fig:Motivation}-B).
W.r.t this issue, we propose a novel scheme that is capable of emphasizing the semantic information usage further during the image SOD procedure.

\begin{figure*}[!t]
\centering
\includegraphics[width=1\linewidth]{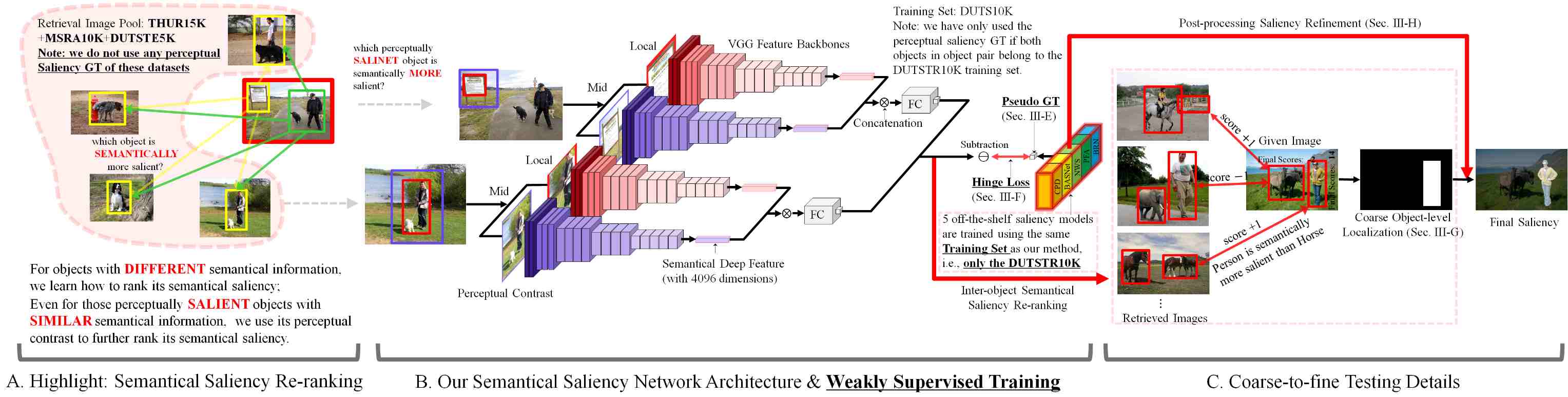}
\caption{Method Overview. The sub-figure A demonstrates our learning objectives; because the high-level semantic information rooted in our brain plays an extremely important role to determine where we look at, we attempts to learn the object-level semantic ranks between a sub-group similar images; The sub-figure B shows the network details, which is lightweight designed, weakly supervised, with fixed backbones, aiming to locate those semantically salient regions coarsely; The sub-figure C shows how to use multiple off-the-shelf deep models to perform pixel-wise saliency refinement.}
\label{fig:Pipeline}
\end{figure*}

Different to the conventional methods~\cite{ECCV16Lu} which focus on learning the perceptual saliency in a fully supervised manner, we devise a novel scheme to learn the object-level semantic saliency ranks in a weakly supervised fashion.
We show the method overview in Fig.~\ref{fig:Pipeline}.
Our semantic saliency network also consists of two sequential sub-nets, i.e., one feature backbone network followed by a lightweight binary classification network.
Our network simultaneously takes two resized rectangular object proposals as input, whose backbone sub-net aims to compute the semantic deep features for each of them.
Then, for any two objects with different semantic information, the learning objective of our classification sub-net is quite simple and intuitive, attempting to make a binary decision on which input object is semantically more salient.
In the case of two objects with similar semantic information, we use the object-level perceptual contrast to alleviate the learning ambiguity.
Once our semantic saliency network has been trained, we directly use it to conduct the inter-object semantic saliency ranking, aiming to coarsely locate those potentially salient regions as demonstrated in the right part of Fig.~\ref{fig:Pipeline}.
Then, as a post-processing procedure, we selectively fuse multiple off-the-shelf deep saliency models to achieve the pixel-wise saliency refinement for the final saliency.

It also should be noted that the conventional methods learning perceptual saliency and conducting pixel-wise refinement at the same time easily lead to an extremely large problem domain, thus their feature backbones need to be completely re-trained to ensure its learning convergency, yet such re-training procedure easily degenerates its feature backbone in revealing semantic information.
In sharp contrast, the problem domain of our semantic saliency learning is much more simple, which focuses on ranking the object-level semantic saliency solely, enabling to fix its feature backbone during training, preserving more semantic information, and ensuring high quality SOD eventually.

\section{Related Work}
Over the past decades, many image SOD methods have been developed~\cite{ChenPR16,OurTIP15}, and these methods can be roughly divided into two categories: the conventional methods using handcrafted features~\cite{yang2013saliency,cheng2014global,harel2007graph,xie2012bayesian,cheng2013efficient,liu2010learning,borji2012exploiting}  and the current main stream methods using deep learning techniques~\cite{li2016visual,liu2016dhsnet,wang2015deep,chen2017video,HouPami19Dss,wang2017deep}.

Long et al.~\cite{long2015fully} first proposed a fully convolutional network to predict the saliency label of each pixel, more and more pixel-level saliency detection methods have been proposed.
Hou et al.~\cite{HouPami19Dss} proposed a short connection to integrate the high-level semantic information of the deep layers into the shallower layers, taking full use of the multi-scale deep features, improving the overall detection performance significantly.
Then, to further make use of the multi-scale deep features, Zhang et al.~\cite{zhang2017amulet} combined multi-level resolution feature maps by bidirectional aggregation method, which integrate the multi-scale information in both the top-down and the bottom-up ways.

Many deep learning models that make full use of feature integration are proposed.
Wang et al.~\cite{wang2017stagewise} proposed a pyramid pool module and a multi-stage refinement mechanism to collect context information and stage-wise results.
Luo et al.~\cite{luo2017non} proposed a simplified convolutional neural network that combines global information and local information through a multi-resolution grid structure.
Zhang et al.~\cite{zhang2017learning} utilized the deep uncertain convolutional features and proposed a reformulated dropout after specific convolutional layers to construct an uncertain ensemble of internal feature units.
Li et al.~\cite{chen2018progressively} proposed a complementary perceptual network that combines cross-over models and cross-layer features to solve the saliency detection task of depth information.
However, they do not typically consider the object-level semantic saliency information.

Our method is a weakly supervised object-level semantic saliency detection scheme compared with the conventional methods~\cite{li2016visual,li2016deep,wang2015deep,wang2017deep} that focuses on learning the perceptual saliency with a fully supervised method.
Moreover, our approach is also totally different to the co-saliency detection approaches~\cite{fu2013cluster,zhang2019co,tsai2018image,CongTIP18}.
Co-saliency detection aims at detecting the common perceptual saliency consistency from multiple images, while our approach mainly resorts the semantic saliency to coarsely locate those salient objects.

Siamese network~\cite{chopra2005learning} is a network with two identical net-work branches and a loss module.
The two branches share weights during training.
Pairs of images and labels are the input of the network, yielding two outputs which are passed to the loss module.
The gradients of the loss function with respect to all model parameters are computed by back propagation and updated with the stochastic gradient method.

\section{Proposed Method}
Our method follows a coarse-to-fine manner, in which the object-level semantic ranking part aims to coarsely locate those semantically salient regions via rectangular boxes.
Here, we will introduce it in details.

\subsection{Object Proposal Preliminary}
\label{sec:OPP}
Since our network attempts to learn the ``object-level'' semantic saliency, we use the off-the-shelf GARPN~\cite{wang2019region} to decompose the given image into 10 rectangular object proposals (it can be more than 10 proposals, but we empirically choose to use 10 proposals for efficiency).
In order to filter those overlapped object proposals, we compute the intersection over union (IOU) rate~\cite{rahman2016optimizing} between each 2 of 10 object proposals, excluding those object proposals with a smaller confidence degree.
Given an image $I$ and retrieved images $I^{\dotplus}$, we only consider object pairs $\{ I_i,I_{j}^{\dotplus} \}$ and drop the $\{ I_i^{\dotplus},I_{j}^{\dotplus} \}$ cases.
If the object covers a similar rectangular area, we will drop the smaller one, because we tend to keep high Recall during localization, and the decreased Precision will be compensated via the latter pixel-wise refinement.
Consequently, each object in the given image can be efficiently and accurately warped by a single rectangle box tightly, and the rectangular proposal will be resized to a fixed size to fit the adopted feature backbone.

\subsection{Object-level Semantic Deep Feature}
\label{sec:OSD}
The overall network architecture of our method is quite simple, which simultaneously takes 2 resized rectangular object proposals as input.
For each input object proposal with size $w\times h$, we use the off-the-shelf backbone network (VGG16) to compute its high dimensional semantic deep feature (4096).

To ensure a high discriminative deep feature space, for each object proposal ($P$), we further compute the deep feature of its enlarged version ($\hat{P}$, with size $\{1.5\times w\}\times\{1.5\times h\}$) via backbone network ($Backbone$) to comprise more nearby surroundings, which will then be concatenated with the original 4096 feature as the 4096+4096 multi-scale perceptual contrast (Local+Mid, $f$), see Eq.~\ref{eq:multiscalef}.
\begin{equation}
f \gets Backbone(P)\otimes Backbone(\hat{P}).
\label{eq:multiscalef}
\end{equation}

\subsection{Network Architecture}
\label{sec:NA}
Since our network aims to coarsely locate those semantically salient regions, there are two feasible choices to design our semantic sub-net:\\
1) A typical choice is to directly feed the object-level semantic deep features into a full connected sub-net (we have tested the FC6) to make a binary prediction of whether the input object proposal is semantically salient or not;\\
2) Alternatively, we can also design the siamese semantic sub-net to receive 2 semantic deep features (from different object proposals) at each time, aiming to rank the inter-object semantic saliency, predicting which one is semantically more salient.

In fact, it is quite ubiquitous that two semantically similar objects are totally different in saliency degrees, which leads the choice 1) to the learning ambiguity.
Thus, the learning objective of choice 1) seems to be problematically formulated.
In sharp contrast, the choice 2) has reformulated the semantic saliency learning problem into the semantic saliency re-ranking problem, and thus it is more reasonable than choice 1).
Also, the learning objective of choice 2) is quite simple and intuitive, which aims to enlarge the semantic deep feature distance between two objects even in the case that these two objects are semantically similar.
Moreover, the quantitative comparisons (Table.~\ref{table:DifferentSubnet}) between choice 1) and 2) also suggest to adopt the latter option.
Therefore, we implement our semantic sub-net by using the choice 2), and the network architecture overview can be found in Fig.~\ref{fig:Pipeline}.

\begin{figure}[!t]
\centering
\includegraphics[width=1\columnwidth]{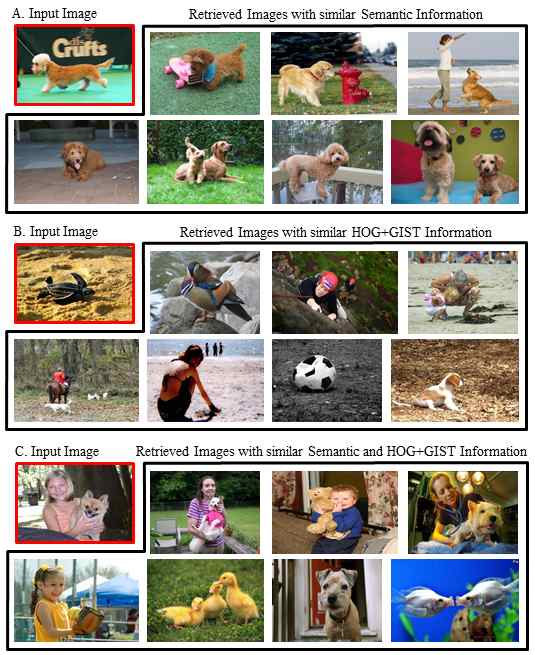}
\caption{Different retrieval schemes. The hybrid image retrieval scheme (C) ensures that the object semantic categories in images with either similar semantic information or similar overall scene layout tend to reach a stable co-occurrence state for its objects. Thus, the hybrid retrieval scheme is more suitable for our object-level semantic saliency re-ranking problem.}
\label{fig:ReDemo}
\end{figure}

\subsection{Object-level Training Data}
\label{sec:TDP}
Since our method aims to rank the inter-object semantic saliency, the learning scope should comprise the following cases: 1) semantically ``different'' object pairs; 2) semantically ``similar'' object pairs.
We formulate the training set to include both the intra-image and inter-image object pairs, seeing the pictorial demonstrations in left part of Fig.~\ref{fig:Pipeline}.
By using the aforementioned object proposals extraction method, we obtain multiple non-overlapped object proposals (max. 10) from each image in our image pool (THUR15K\cite{ChengGroupSaliency}$+$MSRA10K\cite{cheng2014global}$+$DUTSTE5K\cite{zhao2015saliency}), and each two of these objects will be composed as one intra-image training instance.

Meanwhile, for each image in our training set DUTSTR10K, we use it to retrieve $K$ similar images from image pool\{THUR15K$+$MSRA10K$+$DUTSTE5K\} as the retrieved sub-group images, and two object proposals from different images in this sub-group will be composed as an inter-image training instance.
Also, for images with either similar semantical information or similar overall scene layout, its object-level semantic categories tend reach a stable co-occurrence status, e.g., a \emph{car} frequently accompanies with a \emph{road}.
Thus, we adopt the hybrid image retrieval scheme (see Fig.~\ref{fig:ReDemo}), i.e., the sub-group $K$ images are composed by $K_s$ semantically similar images (images with top-$K_s$ of the descending order of the image-level VGG features distance) and $K_l$ images with similar scenes (using GIST+HOG), where $K=K_s+K_l$.
In our implementation, we assign $K=5$, $K_s=2$, $K_l=3$ as the optimal choice according to our component evaluation.

\subsection{Semantic Saliency Pseudo-GT}
\label{sec:SSP}
Thus far, we have converted the original training set into object pairs.
For each object pair ($\{P_1,P_2\}$), our network aims to make a binary decision on whether the object $P_1$ is semantically more salient than the object $P_2$.
For each object $P$ in the training set (DUTSTR10K), we assign its perceptual saliency label ($PSL$) = 1 if the rectangular overlap rate between this object and its perceptual saliency GT map (provided by the training set) $>70\%$, and assign the $PSL = -1$ otherwise.

Thus, for an object pair ($\{P_1,P_2\}$) (both objects belong to the training set DUTSTR10K) with different perceptual saliency labels, we formulate its semantic saliency pseudo-GT ($pGT$) as Eq.~\ref{eq:pgt1}.
\begin{equation}
\label{eq:pgt1}
pGT_{\{P_1,P_2\}}=\left\{\begin{array}{ll} \ \ 1\ \ if\ PSL_{P_1}>PSL_{P_2}\\ \\-1\ \   otherwise\end{array}\right.,
\end{equation}
In the case of object pair ($\{P_1,P_2\}$) with identical perceptual saliency labels (both objects belong to the training set DUTSTE10K) or one of its objects belongs to the image pool (THUR15K+MSRA10K+DUTSTE5K), we use 5 off-the-shelf deep saliency models, which are all pre-trained using the \textbf{Same Training Set} as our method (i.e., the DUTSTE10K only), to make its semantic saliency pseudo-GT ($pGT$) as Eq.~\ref{eq:pgt2}.
\begin{equation}
\label{eq:pgt2}
pGT_{\{P_1,P_2\}}=\left\{\begin{array}{ll} \ \ 1\ \ if\ PSL5_{P_1}>PSL5_{P_2}\\ \\-1\ \   otherwise\end{array}\right.,
\end{equation}
where $PSL5_{P_1}=||pat(S, P_1)||_1/(w_1\cdot h_1)$; $w_1,h_1$ respectively denote the width and height of the object proposal $P_1$; the $S=\sum Sal_i, i\in\{1,...,5\}$, $Sal_i$ denotes the saliency map predicted by the $i$-th pre-trained saliency model; $pat(S,P_1)$ returns the $P_1$ correlated patch in $S$; $||\cdot||_1$ is the $L_1$-norm.
Our rationale is to give an object pair (no matter whether these 2 objects are semantically similar or not), and the object with a higher saliency prediction from the 5 pre-trained deep models is perceptually more salient than the other one, which suggests that its semantic information should be more salient.

\begin{figure}[t]
\centering
\includegraphics[width=1\columnwidth]{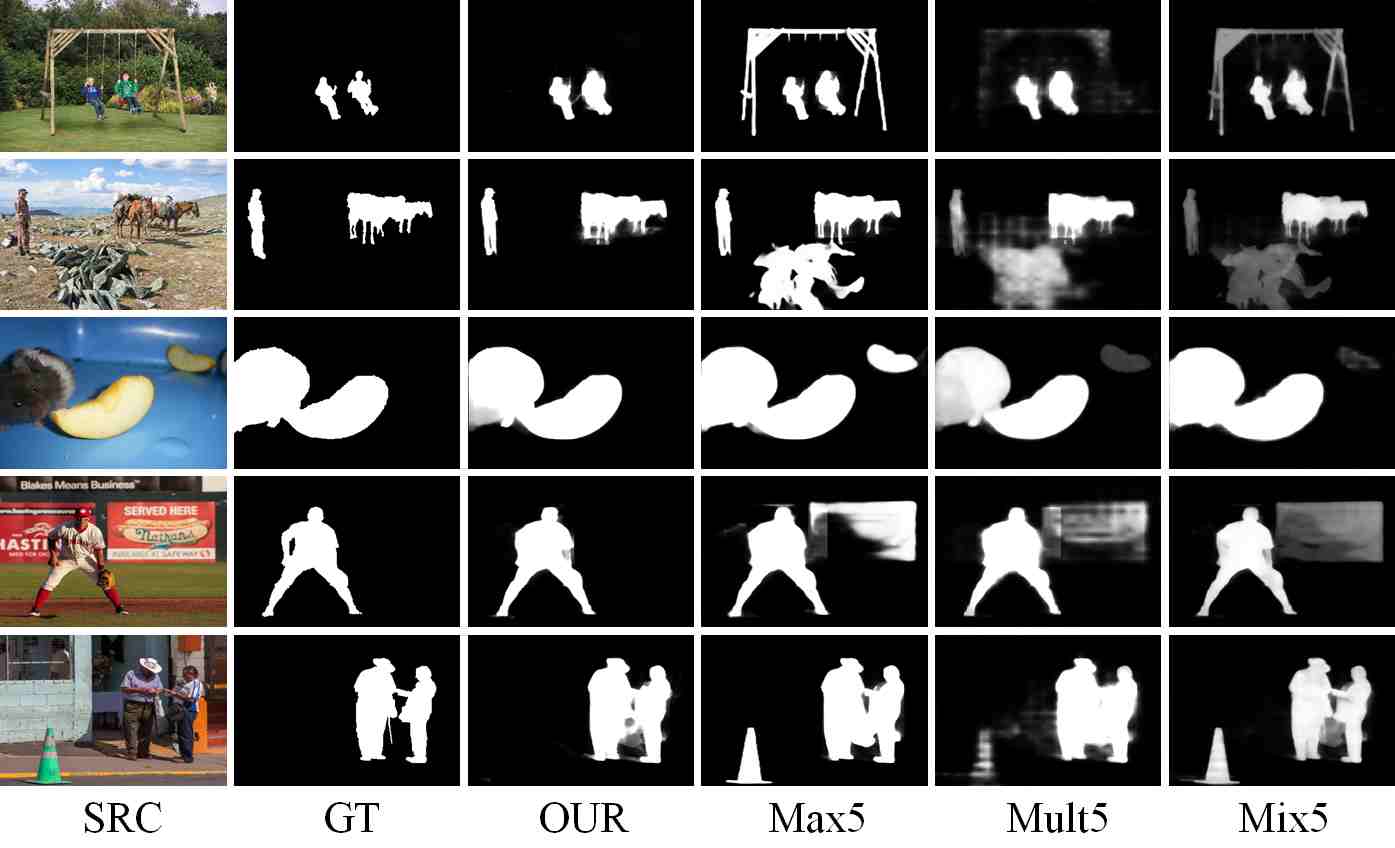}
\caption{Visual comparisons between our method and multiple classic fusion schemes; Max5, Mult5, and Mix5 respectively represent the saliency maps by simply fusing the 5 adopted deep models via maximizing, multiplicative and average; the adopted 5 deep models include CPD~\cite{wu2019cascaded}, BASNet~\cite{Qin_2019_CVPR}, MWS~\cite{zeng2019multi}, PFA~\cite{zhao2019pyramid} and BRN~\cite{wang2018detect}.}
\label{fig:ComponentDemo}
\end{figure}

\subsection{Loss Function}
Thus far, our object-level training set can be represented as $X=\{f_1,f_2\},Y=pGT$, where X and Y respectively denote the training instance (i.e., object pair deep features) and its binary label (Eq.~\ref{eq:pgt1} and Eq.~\ref{eq:pgt2}).
Since the binary label $pGT$ is formulated in a weakly supervised manner, the $pGT$ may occasionally conflict with the real semantic rank between $f_1$ and $f_2$. Thus, we choose to use the hinge loss (Eq.~\ref{eq:loss}).
\begin{equation}
\label{eq:loss}
L(f_1,f_2,\theta) = max\Bigg\{0, pGT\cdot\Big(FC_2(f_2)-FC_1(f_1)\Big)-\rho\Bigg\},
\end{equation}
where $\theta$ denotes the network parameters, $FC_1$ and $FC_2$ are two full connection sub branches (with size 8192-1024-2048-2048-1024-1024-2) in our siamese network (see Fig.~\ref{fig:Pipeline}); $\rho$ is the hinge loss margin (we empirically assign it to 10) which vanishes the gradient to alleviate the learning ambiguity in the case of two objects with almost the same semantic saliency, alleviating the learning ambiguity.
The hinge loss is able to focus its gradient on the semantic saliency ranking problem, which is more suitable than the widely used entropy loss, and the quantitative proofs can be found in Table.~\ref{table:DifferentLoss}.

\subsection{Semantically Salient Object Localization}
\label{sec:SSOL}
Given a testing image $I$, here, we aim to coarsely determine which objects in this image are semantically salient.
For each object ($P_i$) in $I$, we define its semantic saliency degree (Score) as Eq.~\ref{eq:SSD}.
\begin{equation}
\label{eq:SSD}
Score_i=\sum_{P_j\in I^+ or I} \Bigg|\Bigg|\max\Big\{0,FC_1(f_i)-FC_2(f_j)\Big\}\Bigg|\Bigg|_0,
\end{equation}
where $||\cdot||_0$ is the $L_0$-norm; $f_i$ denotes the deep feature of the $i$-th object in image $I$; $FC_1$ and $FC_2$ are the learned sub branches (shared weights) in our siamese network, which are identical to Eq.~\ref{eq:loss}; $I^+$ denotes the sub group images with $K$ similar images to $I$ (Sec.~\ref{sec:TDP}).

By using Eq.~\ref{eq:SSD}, those semantically salient objects in $I$ will be assigned with large $Score$. However, we do not know the exact number of how many objects in $I$ are the semantically salient ones, which hinders us to directly use $Score$ value for the localization of semantically salient objects.

Thus, we select the top-$q$ objects with largest $Score$ in $I$ as those semantically salient ones, where the value of $q$ is adaptively determined by Eq.~\ref{eq:q}.
\begin{equation}
\label{eq:q}
q=arg\max\limits_i\{\frac{\partial \xi}{\partial i}\},\ \  \xi\gets des\Big([Score_1, Score_2, ...]\Big),
\end{equation}
where $\partial$ denotes the partial derivative; $des(\cdot)$ ranks its input according to the $Score$ values in descending order.
The rationale of Eq.~\ref{eq:q} is that those semantically salient objects in $I$ should simultaneously have the following attributes: 1) with an extremely large $Score$; 2) and its $Score$ value should also be significantly larger than those of the semantically non-salient objects.
The semantically salient objects determined by our scheme are more accurate than the conventional models to locate the true salient object, seeing the quantitative proofs in Table.~\ref{table:DifferentAq}.
In practice, our method can allow “multiple” objects to go the saliency refinements if their semantic scores are similarly high (see Eq..~\ref{eq:SSD} and Eq.~\ref{eq:q}). Thus, our method can perform well for datasets with multiple salient objects belonging to different categories.

\begin{figure*}[t]
\centering
\includegraphics[width=1\linewidth]{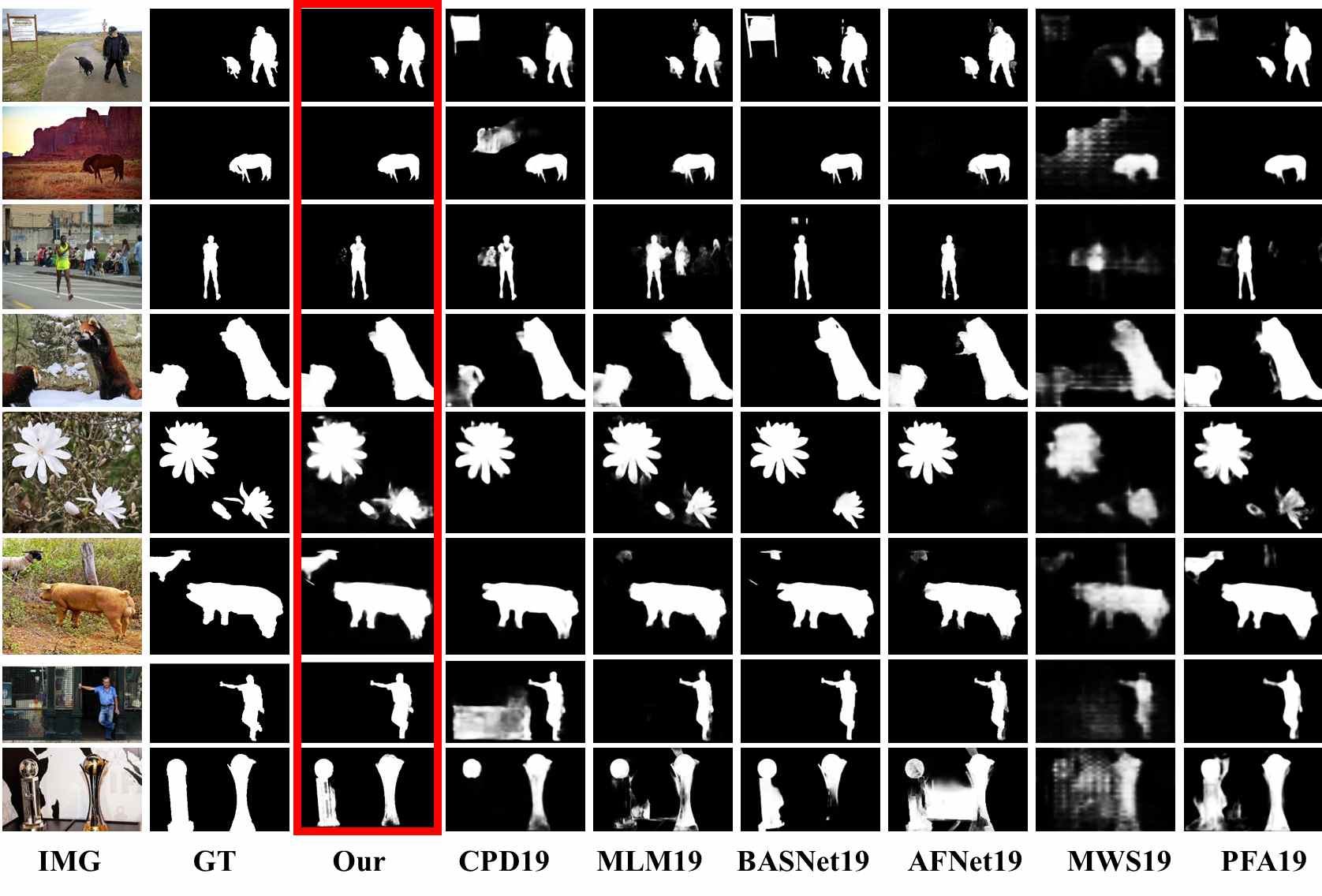}
\caption{Qualitative comparisons between our method and several most recent deep models, including CPD19~\cite{wu2019cascaded}, MLM19~\cite{wu2019mutual}, BANet19~\cite{su2019banet}, AFNet19~\cite{Feng2019CVPR}, MWS19~\cite{zeng2019multi} and PFA19~\cite{zhao2019pyramid}.}
\label{fig:DifferentSaliencyMaps}
\end{figure*}

\subsection{Pixel-wise Saliency Refinement}
\label{sec:PSR}
Thus far, we have coarsely located $q$ semantically salient objects in the given image $I$, seeing the white rectangular in the right part of Fig.~\ref{fig:Pipeline} (we denote it as $I_c$).
To highlight the real salient object while compressing its non-salient nearby surroundings, we use 5 off-the-shelf deep saliency models (we use $Sal_i, i\in\{1,2,3,4,5\}$ to represent their saliency predictions), which are identical to Sec.~\ref{sec:SSP} using the same training set as our method, to conduct the pixel-wise saliency refinement as a post-processing procedure.

In general, the saliency predictions made by the 5 off-the-shelf models may vary one another, and it is not advisable to directly fuse 5 saliency maps into the $q$ coarsely localized rectangular regions (i.e., $I_c$).
Since the $q$ localized object boxes are trustworthy to locate those potentially salient regions, we use the overlap rate between these $q$ boxes and each of the 5 saliency maps to measure the confidence degree of the adopted 5 off-the-shelf saliency models.
Thus, the confidence degree of the $i$-th deep saliency model for the $j$-th object box in $I$ can be formulated as Eq.~\ref{eq:conf}:
\begin{equation}
\label{eq:conf}
\begin{split}
Conf(i,j)=\underset{local\ similarity} {\underbrace{\Bigg|\Bigg|\frac{pat(Sal_i,P_j)}{pat(I_c,P_j)}\Bigg|\Bigg|_1}} + \lambda\cdot \underset{global\ similarity} {\underbrace{\Bigg|\Bigg|\frac{Sal_i\odot I_c+C}{Sal_i+I_c+C}\Bigg|\Bigg|_1}},
\end{split}
\end{equation}
where $C$ is a small constant value to avoid division by zero; $\odot$ denotes the element-wise Hadamard product; function $pat(Sal_i,P_j)$ returns the $P_j$ correlated patch in $Sal_i$; $\lambda$ is a balance factor between the local and global similarity, and we empirically assign it to 0.5.

The final saliency map ($FinalS$) can be computed by adaptively fusing the saliency maps of 5 deep saliency models as Eq.~\ref{eq:fs}.
\begin{equation}
\label{eq:fs}
FinalS = \sum\limits_{i=1}^5\sum_{j=1}^q Conf(i,j)\cdot \Big\{M(I_c,P_j)\odot Sal_i\Big\},
\end{equation}
where function $M(I_c,P_j)$ returns a binary mask matrix $Mask\in\{0,1\}$ (with identical size to $I_c$), in which only those pixels in object box $P_j$ are non-zero elements.
The saliency map quality computed by our adaptive fusion scheme can significantly outperform the conventional fusion scheme, seeing the quantitative proofs in Table.~\ref{table:DifferentAq} and Table.~\ref{table:DifferentCombine}.

\subsection{Why Our Method is a Weakly-supervised One?}
The main reason is that we do not directly use the saliency outputs of the adopted 5 pre-trained methods in our training. Instead, our pseudo-GT is just the saliency ranks, which are newly formulated by measuring the IOU rate of the five adopted models.
So, it is appropriate to regard our method as weakly supervised.

The novelty of our weakly supervised learning can be
confirmed by the following 2 facts:
\textbf{1)} The main purpose is different, in which our method aims
to investigate the object-level semantic ranks between similar images,
whereas the previous works were mainly designed for pixel-wise saliency
regression in single image;
\textbf{2)} The key methodology is different, in which we have adopted
the object-level IOU rate to formulate the semantic pseudo-GT, making
the key idea of using semantic ranking for SOD feasible.

\subsection{Major Differences to the SOTA Methods}
The previous works (e.g., CPD~\cite{wu2019cascaded}) have treated the SOD as a multi-task problem
which aims to conduct pixel-wise saliency regression and segmentation-like
saliency refinement simultaneously, making their networks difficult to
reach convergence.
Consequently, even at the expense of decreasing semantic information usage,
the conventional methods must update their feature backbones (such as the
vanilla VGG16) to ensure learning convergence.
In sharp contrast, our method has divided the SOD problem into 2 sequential tasks:
localization first and refinement latter.
Since our localization has a relatively small problem domain, it can be fully
realized from the semantic perspective by using ``fixed'' backbones
with shallower FC layers.
Moreover, different from the conventional methods which focused on
learning pixel-wise saliency in ``single image using perceptual clues mainly'',
our method has investigated the ``object-level semantic ranks between
multiple images'', which is more consistent with both human vision system
and human brain itself.

\section{Experiments and Results}
We have conducted massive quantitative experiments to validate the effectiveness of our method. We also have compared our method with 11 SOTA methods over 5 public available datasets to demonstrate the advantages of our method.
The testing datasets adopted in our quantitative evaluation include DUTS-TE~\cite{zhao2015saliency}, ECSSD~\cite{yan2013hierarchical}, HKU-IS~\cite{li2016deep}, PASCAL-S~\cite{li2014secrets} and SOD~\cite{movahedi2010design}.

\subsection{Evaluation Metrics}
We adopt the F-measure and the mean absolute error (MAE) to evaluate the performance of our method, including: the F-measure and the mean absolute error (MAE).
As the recall rate is inversely proportional to the precision, the tendency of the trade-off between precision and recall can truly indicate the overall saliency detection performance.
Thus, we utilize the F-measure ($\beta^2$=0.3) to evaluate such trade-off.
Moreover, since both metrics of MAE and F-measure are based on pixel-wise errors and often ignore the structural similarities, we also adopt the structure measure S-measure~\cite{fan2017structure} and  enhanced-measure E-measure~\cite{fan2018enhanced} to conduct quantitative evaluation.
The evaluation code source code is provided by Fan et al.~\cite{fan2019SOC,fan2017structure,fan2018enhanced}.

\subsection{Training and Implementation Details}
We implement our approach in Python with the Pytorch toolbox. We run our approach on a computer with 4 Tesla P100 GPU (with 64G memory).
The retrieval image pool consists of almost 30K images, i.e., THUR15K\cite{ChengGroupSaliency}$+$MSRA10K\cite{cheng2014global}$+$DUTSTE5K\cite{zhao2015saliency}.
It should be noted that we do not use any perceptual saliency ground truth of these 30K images.
Our training set only comprises 10K images from the DUTSTR10K~\cite{zhao2015saliency} dataset, and we have applied its perceptual saliency ground truth to facilitate the pseudo-GT formulation for the object pairs (30\%) with all its objects belonging to the training set.
As for the rest of 70\% object pairs, we weakly formulate their pseudo-GT by using the saliency predictions of 5 off-the-shelf models.
To avoid using testing data, the adopted 5 off-the-shelf models are all pre-trained using the same training set as our method (i.e., DUTSTR10K), and the selected 5 methods include CPD~\cite{wu2019cascaded}, BASNet~\cite{Qin_2019_CVPR},MWS~\cite{zeng2019multi},PFA~\cite{zhao2019pyramid} and BRN~\cite{wang2018detect}.

\begin{table*}[t]
\caption{Quantitative comparisons between our method and SOTA methods in metrics including maximum F-measure (larger is better), MAE (smaller is better) , E-measure (larger is better), adaptive F-measure (larger is better) and adaptive E-measure (larger is better).  The top three results are highlighted in {\textcolor{red}{red}}, {\textcolor{green}{green}}, and {\textcolor{blue}{blue}}, respectively.}
\vspace{0.1cm}
\renewcommand\arraystretch{1.1}
  \resizebox{1\linewidth}{!}{
\begin{tabular}{|p{0.2cm} | p{1.8cm} | p{0.8cm}<{\centering} p{0.8cm}<{\centering} p{0.8cm}<{\centering}  p{0.8cm}<{\centering}  p{0.8cm}<{\centering}  p{0.8cm}<{\centering} p{0.8cm}<{\centering} p{0.8cm}<{\centering}  p{0.8cm}<{\centering}  p{0.8cm}<{\centering}  p{0.8cm}<{\centering}  p{0.8cm}<{\centering} | p{0.8cm}<{\centering} | p{0.8cm}<{\centering}  | p{0.8cm}<{\centering}| p{0.8cm}<{\centering}| }
\bottomrule
\hline

\multirow{2}{*}{}& \multirow{2}{*}{Metric}&OUR&EGNet&BANet&AFNet &CPD & MLM& BASNet &MWS & PFA & BRN & RFCN & RADF \\
& &&19~\cite{zhao2019egnet}&19~\cite{su2019banet}&19~\cite{Feng2019CVPR} &19~\cite{wu2019cascaded} & 19~\cite{wu2019mutual} & 19~\cite{Qin_2019_CVPR}& 19~\cite{zeng2019multi} & 19~\cite{zhao2019pyramid} & 18~\cite{wang2018detect} & 18~\cite{wang2018salient} &18~\cite{hu2018recurrently} \\
\hline
\hline
\multirow{8}{*}{\rotatebox{90}{DUTS-TE~\cite{zhao2015saliency}}}
&Smeasure$\uparrow$ &{\textcolor{red}{.909}}&{\textcolor{green}{.887}}&{\textcolor{blue}{.879}}&.867&.869&.862&.866&.759&.874&.851&.792&.824 \\
&MAE$\downarrow$    &{\textcolor{red}{.039}}&{\textcolor{red}{.039}}&{\textcolor{green}{.040}}&.046&.043&.049&.048&.091&{\textcolor{blue}{.041}}&.054&.074&.072 \\
&adpEm$\uparrow$    &{\textcolor{red}{.928}}&{\textcolor{blue}{.891}}&{\textcolor{green}{.892}}&.879&.886&.860&.884&.814&.877&.860&.837&.819 \\
&meanEm$\uparrow$   &.894&{\textcolor{blue}{.907}}&{\textcolor{red}{.913}}&.893&.898&.883&.895&.743&{\textcolor{green}{.910}}&.876&.826&.840 \\
&maxEm$\uparrow$    &{\textcolor{red}{.962}}&{\textcolor{blue}{.927}}&{\textcolor{blue}{.927}}&.910&.914&.907&.903&.833&{\textcolor{green}{.933}}&.899&.854&.874 \\
&adpFm$\uparrow$    &{\textcolor{red}{.839}}&{\textcolor{green}{.815}}&{\textcolor{green}{.815}}&.792&{\textcolor{blue}{.805}}&.745&.791&.684&.784&.755&.709&.700 \\
&meanFm$\uparrow$   &{\textcolor{red}{.841}}&{\textcolor{green}{.839}}&{\textcolor{blue}{.835}}&.812&.821&.792&.822&.648&.815&.780&.713&.737 \\
&maxFm$\uparrow$    &{\textcolor{red}{.910}}&{\textcolor{green}{.866}}&{\textcolor{blue}{.858}}&.838&.840&.827&.838&.720&.852&.811&.736&.786 \\

\hline
\multirow{8}{*}{\rotatebox{90}{ECSSD~\cite{yan2013hierarchical}}}
&Smeasure$\uparrow$ &.836&{\textcolor{red}{.925}}&{\textcolor{green}{.924}}&.913&.910 &.911 &{\textcolor{blue}{.917}} &.828 &.903 &.909 &.869 &.895 \\
&MAE$\downarrow$    &.044&.{\textcolor{red}{037}}&{\textcolor{green}{.035}}&.042&{\textcolor{blue}{.040}} &.045 &{\textcolor{red}{.037}} &.096 &.046 &.046 &.067 &.060 \\
&adpEm$\uparrow$    &{\textcolor{red}{.934}}&{\textcolor{blue}{.927}}&{\textcolor{green}{.928}}&.918&.922 &.914 &.922 &.885 &.910 &.915 &.907 &.908 \\
&meanEm$\uparrow$   &.918&{\textcolor{blue}{.943}}&{\textcolor{red}{.948}}&.935&.938 &.927 &{\textcolor{green}{.944}} &.792 &.936 &.930 &.897 &.907 \\
&maxEm$\uparrow$    &{\textcolor{red}{.972}}&{\textcolor{blue}{.955}}&{\textcolor{green}{.958}}&.947&.944 &.945 &.952 &.910 &.948 &.947 &.921 &.933 \\
&adpFm$\uparrow$    &.909&{\textcolor{green}{.920}}&{\textcolor{red}{.923}}&.908&{\textcolor{blue}{.915}} &.869 &.882 &.840 &.886 &.893 &.871 &.872 \\
&meanFm$\uparrow$   &.899&{\textcolor{green}{.918}}&{\textcolor{red}{.923}}&.905&.912 &.890 &{\textcolor{blue}{.917}} &.763 &.889 &.893 &.866 &.893 \\
&maxFm$\uparrow$    &{\textcolor{red}{.955}}&{\textcolor{blue}{.936}}&{\textcolor{green}{.939}}&.924&.923 &.918 &.931 &.859 &.913 &.917 &.885 &.905 \\

\hline
\multirow{8}{*}{\rotatebox{90}{HKU-IS~\cite{li2016deep}}}
&Smeasure$\uparrow$ &{\textcolor{red}{.930}}&{\textcolor{green}{.918}}&{\textcolor{blue}{.913}}&.905&.905&.906&.908&.818&{\textcolor{blue}{.913}}&.901&.862 &.888  \\	
&MAE$\downarrow$    &.037&{\textcolor{red}{.031}}&{\textcolor{green}{.032}}&.036&.034&.039&{\textcolor{green}{.032}}&.084&{\textcolor{blue}{.033}}&.041&.054 &.050  \\	
&adpEm$\uparrow$    &{\textcolor{red}{.962}}&{\textcolor{green}{.950}}&{\textcolor{green}{.950}}&.942&{\textcolor{blue}{.944}}&.937&.945&.895&.946&.939&.924 &.919  \\	
&meanEm $\uparrow$  &.916&{\textcolor{blue}{.944}}&{\textcolor{green}{.946}}&.934&.938&.930&.943&.787&{\textcolor{red}{.948}}&.929&.897 &.909  \\	
&maxEm  $\uparrow$  &{\textcolor{red}{.974}}&{\textcolor{blue}{.958}}&{\textcolor{blue}{.958}}&.949&.950&.950&.951&.908&{\textcolor{green}{.962}}&.949&.931 &.938  \\
&adpFm  $\uparrow$  &{\textcolor{red}{.912}}&{\textcolor{green}{.901}}&{\textcolor{blue}{.900}}&.888&.890&.871&.895&.814&.883&.875&.858 &.851  \\	
&meanFm $\uparrow$  &.884&{\textcolor{green}{.902}}&{\textcolor{red}{.903}}&{\textcolor{blue}{.888}}&.891&.878&{\textcolor{green}{.902}}&.734&.891&.875&.848 &.856  \\	
&maxFm  $\uparrow$  &{\textcolor{red}{.943}}&{\textcolor{green}{.924}}&{\textcolor{blue}{.923}}&.910&.910&.910&.918&.835&.918&.903&.872 &.895  \\

\hline
\multirow{8}{*}{\rotatebox{90}{PASCAL-S~\cite{li2014secrets}}}
&Smeasure$\uparrow$ &{\textcolor{red}{.894}}&{\textcolor{blue}{.849}}&{\textcolor{blue}{.849}}&{\textcolor{blue}{.849}}&.846 &.841 &.835 &.765 &{\textcolor{green}{.859}} &.841 &.796  &.815 \\
&MAE$\downarrow$    &{\textcolor{red}{.065}}&.076&.072&{\textcolor{blue}{.070}}&.072 &.076 &.078 &.135 &{\textcolor{green}{.066}} &.079 &.105  &.101 \\
&adpEm$\uparrow$    &{\textcolor{red}{.894}}&.852&{\textcolor{green}{.857}}&.851&{\textcolor{blue}{.853}} &.840 &.850 &.789 &.840 &.838 &.830  &.821 \\
&meanEm$\uparrow$   &.879&.879&{\textcolor{green}{.889}}&{\textcolor{blue}{.883}}&.880 &.873 &.876 &.732 &{\textcolor{red}{.894}} &.871 &.828  &.832 \\
&maxEm$\uparrow$    &{\textcolor{red}{.943}}&.889&{\textcolor{blue}{.897}}&.895&.889 &.889 &.883 &.830 &{\textcolor{green}{.917}} &.888 &.847  &.854 \\
&adpFm$\uparrow$    &{\textcolor{red}{.846}}&.821&{\textcolor{green}{.828}}&.822&{\textcolor{blue}{.825}} &.762 &.773 &.716 &.818 &.798 &.772  &.765 \\
&meanFm$\uparrow$   &{\textcolor{red}{.845}}&{\textcolor{blue}{.826}}&{\textcolor{green}{.834}}&{\textcolor{blue}{.826}}&.823 &.805 &.820 &.670 &.823 &.807 &.772  &.775 \\
&maxFm$\uparrow$    &{\textcolor{red}{.902}}&.844&{\textcolor{blue}{.852}}&.845&.837 &.833 &.837 &.756 &{\textcolor{green}{.858}} &.833 &.788  &.808 \\

\hline
\multirow{8}{*}{\rotatebox{90}{SOD~\cite{movahedi2010design}}}
&Smeasure$\uparrow$ &{\textcolor{red}{.806}}&{\textcolor{blue}{.802}}&{\textcolor{blue}{.788}}&/&.767&.786&.769&.700 &/ &.779&.719 &.765 \\
&MAE$\downarrow$    &.112&{\textcolor{red}{.099}}&{\textcolor{blue}{.107}}&/&.112&{\textcolor{green}{.108}}&.113&.167 &/ &.109&.146 &.133 \\
&adpEm$\uparrow$    &{\textcolor{red}{.836}}&{\textcolor{green}{.818}}&{\textcolor{blue}{.812}}&/&.793&.800&.777&.775 &/ &.801&.791 &.801 \\
&meanEm$\uparrow$   &.779&{\textcolor{green}{.818}}&{\textcolor{red}{.825}}&/&.778&{\textcolor{blue}{.799}}&.798&.657 &/ &.798&.745 &.783 \\
&maxEm$\uparrow$    &{\textcolor{red}{.912}}&{\textcolor{green}{.868}}&{\textcolor{blue}{.860}}&/&.848&.844&.829&.821 &/ &.843&.820 &.832 \\
&adpFm$\uparrow$    &{\textcolor{blue}{.825}}&{\textcolor{blue}{.840}}&{\textcolor{green}{.831}}&/&.810&.764&.746&.738 &/ &.794&.768 &.776 \\
&meanFm$\uparrow$   &.777&{\textcolor{green}{.819}}&{\textcolor{red}{.822}}&/&.770&.779&{\textcolor{blue}{.790}}&.631 &/ &.782&.737 &.766 \\
&maxFm$\uparrow$    &{\textcolor{red}{.878}}&{\textcolor{green}{.845}}&{\textcolor{blue}{.838}}&/&.814&.806&.805&.772 &/ &.807&.777 &.798 \\

\bottomrule
 \end{tabular}
 }
\label{table:ComparisonDifferentModels}
\end{table*}

\begin{table}[t]
\caption{Quantitative comparisons between different image retrieval choices (we empirically retrieval 5 images), i.e., SE: the semantic information based image retrieval, GH: the classic GIST+HOG image retrieval, HB: the hybrid image retrieval (3 images using SE, and 2 images using GH); ``DUT.'': DUTS-TE~\cite{zhao2015saliency}, ``ECS.'': ECSSD~\cite{yan2013hierarchical}, ``HKU.'': HKU-IS~\cite{li2016deep}, ``PAS.'': PASCAL-S~\cite{li2014secrets}, SOD~\cite{movahedi2010design}; ``Sme.'': Smeasure, ``ad.E'': adpEmeasure, ``me.E'': meanEmeasure, ``ma.E'': maxEmeasure, ``ad.F'': adpFmeasure, ``me.F'': meanFmeasure, ``ma.F'': maxFmeasure.}
\vspace{0.1cm}
\renewcommand\arraystretch{1}
 \resizebox{1\linewidth}{!}{
\begin{tabular}{|c|rcccccccc|}
\bottomrule
{\footnotesize Dataset}& & Sme.&  MAE &  ad.E& me.E& ma.E& ad.F& me.F& ma.F\\
\hline
\hline
\multirow{3}{*}{DUT.}
&\textbf{HB}&.909 &.039 &.928 &.894 &.962 &.839 &.841 &.910\\
&\textbf{SE}&.889 &.038 &.909 &.882 &.918 &.820 &.839 &.847\\
&\textbf{GH}&.864 &.039 &.903 &.889 &.921 &.852 &.839 &.847 \\
\hline
\multirow{3}{*}{ECS.}
&\textbf{HB}&.936 &.044 &.934 &.918 &.972 &.909 &.899 &.955 \\
&\textbf{SE}&.912 &.045 &.923 &.940 &.951 &.928 &.921 &.926 \\
&\textbf{GH}&.869 &.056 &.861 &.873 &.892 &.847 &.825 &.853 \\
\hline
\multirow{3}{*}{HKU.}
&\textbf{HB}&.930 &.037 &.962 &.916 &.974 &.912 &.884 &.943 \\
&\textbf{SE}&.905 &.039 &.947 &.944 &.952 &.915 &.909 &.915 \\
&\textbf{GH}&.911 &.045 &.928 &.917 &.943 &.891 &.862 &.888 \\

\hline
\multirow{3}{*}{PAS.}
&\textbf{HB}&.894 &.065 &.894 &.879 &.943 &.846 &.845 &.902\\
&\textbf{SE}&.867 &.068 &.860 &.888 &.912 &.827 &.835 &.867\\
&\textbf{GH}&.861 &.066 &.850 &.874 &.901 &.821 &.817 &.849 \\

\hline
\multirow{3}{*}{SOD}
&\textbf{HB}&.806 &.112 &.836 &.779 &.912 &.825 &.777 &.878\\
&\textbf{SE}&.758 &.110 &.759 &.772 &.814 &.778 &.765 &.802\\
&\textbf{GH}&.748 &.113 &.738 &.761 &.790 &.756 &.746 &.764 \\
\hline
 \end{tabular}
 }
\label{table:DifferentRetrieval}
\end{table}

\begin{table}[t]
\caption{Quantitative comparison between the single-scale deep feature (i.e., 4096) and multi-scale deep feature (i.e., 4096+4096). ``SS'': Single-scale, ``MS'': Multi-scale, GH: the classic GIST+HOG image retrieval, HB: the hybrid image retrieval (3 images using SE, and 2 images using GH); ``DUT.'': DUTS-TE~\cite{zhao2015saliency}, ``ECS.'': ECSSD~\cite{yan2013hierarchical}, ``HKU.'': HKU-IS~\cite{li2016deep}, ``PAS.'': PASCAL-S~\cite{li2014secrets}, SOD~\cite{movahedi2010design}; ``Sme.'': Smeasure, ``ad.E'': adpEmeasure, ``me.E'': meanEmeasure, ``ma.E'': maxEmeasure, ``ad.F'': adpFmeasure, ``me.F'': meanFmeasure, ``ma.F'': maxFmeasure.}
\vspace{0.1cm}
\renewcommand\arraystretch{1.1}
 \resizebox{1\linewidth}{!}{
\begin{tabular}{|c|rcccccccc|}
\bottomrule

{\footnotesize Dataset}& & Sme.&  MAE &  ad.E& me.E& ma.E& ad.F& me.F& ma.F\\
\hline
\hline
\multirow{2}{*}{DUT.}
& \textbf{SS} &.780  &.616  &.850  &.857  &.863  &.691  &.702  &.711 \\
& \textbf{MS} &.909  &.039  &.928  &.894  &.962  &.839  &.841  &.910  \\

\hline
\multirow{2}{*}{ECS.}
& \textbf{SS} &.699  &.110  &.761  &.766  &.781  &.801  &.819  &.823 \\
& \textbf{MS} &.836  &.044  &.934  &.918  &.972  &.909  &.899  &.955  \\
\hline
\multirow{2}{*}{HKU.}
& \textbf{SS} &.724  &.111  &.766  &.791  &.801  &.752  &.756  &.795 \\
& \textbf{MS} &.930  &.037  &.962  &.916  &.974  &.912  &.884  &.943  \\

\hline
\multirow{2}{*}{PAS.}
& \textbf{SS}&.757  &.129  &.692  &.700  &.702  &.681  &.690  &.714 \\
& \textbf{MS} &.894  &.065  &.894  &.879  &.943  &.846  &.845  &.902  \\

\hline
\multirow{2}{*}{SOD}
& \textbf{SS} &.745  &.161  &.742  &.751  &.761  &.777  &.781  &.788 \\
& \textbf{MS} &.806  &.112  &.836  &.779  &.912  &.825  &.777  &.878  \\
\hline
 \end{tabular}
 }
\label{table:DifferentSubnet}
\end{table}

\subsection{Comparison with SOTA methods}
We have compared our method with other 11 most recent SOTA deep models, which are EGNet~\cite{zhao2019egnet}, BANet~\cite{su2019banet}, AFNet~\cite{Feng2019CVPR}, CPD~\cite{wu2019cascaded}, MLM~\cite{wu2019mutual}, BASNet~\cite{Qin_2019_CVPR}, MWS~\cite{zeng2019multi}, PFA~\cite{zhao2019pyramid}, BRN~\cite{wang2018detect}, RFCN~\cite{wang2018salient}, RADF~\cite{hu2018recurrently}.
For an objective comparison, all quantitative evaluations are conducted using the saliency maps provided by the authors with parameters unchanged.

We demonstrate the detailed S-measure~\cite{fan2017structure}, MAE, F-measure and E-measure~\cite{fan2018enhanced} with different thresholds in Table.~\ref{table:ComparisonDifferentModels}.
And all these quantitative results have indicated the advantages of our method.
For the comparison results over the images with multiple objects (Fig.~\ref{fig:DifferentSaliencyMaps}), the SOTA models occasionally assign large saliency values to those non-salient objects due to their strong perceptual saliency.
However, benefited by our newly learned semantic saliency, our method can solve the above problems effectively.

Specially, as for the quantitative comparisons (Table.~\ref{table:ComparisonDifferentModels}) over the ECCSD dataset, our method may just be able to exhibit a comparable performance to the SOTA methods, because the objects in ECCSD tend to have large difference in semantic to those objects in our retrieval image pool, making our pair-wise object-level semantic saliency difficult. And we believe this problem can be alleviated if we increase the retrieval image pool size or diversity.

\begin{table}[t]
\caption{Quantitative comparison between different network architectures (see details in Sec.~\ref{sec:NA}) using the conventional entropy loss and our hinge loss respectively. ``EL'': Entropy Loss, ``HL'': Hinge Loss, ``DUT.'': DUTS-TE~\cite{zhao2015saliency}, ``ECS.'': ECSSD~\cite{yan2013hierarchical}, ``HKU.'': HKU-IS~\cite{li2016deep}, ``PAS.'': PASCAL-S~\cite{li2014secrets}, SOD~\cite{movahedi2010design}; ``Sme.'': Smeasure, ``ad.E'': adpEmeasure, ``me.E'': meanEmeasure, ``ma.E'': maxEmeasure, ``ad.F'': adpFmeasure, ``me.F'': meanFmeasure, ``ma.F'': maxFmeasure.}
\vspace{0.1cm}
\renewcommand\arraystretch{1.1}
 \resizebox{1\linewidth}{!}{
\begin{tabular}{|c|rcccccccc|}
\bottomrule

{\footnotesize Dataset}& & Sme.&  MAE &  ad.E& me.E& ma.E& ad.F& me.F& ma.F\\
\hline
\hline
\multirow{2}{*}{DUT.}
& \textbf{EL} &.904  &.042  &.910  &.888  &.956  &.869  &.850  &.915 \\
& \textbf{HL} &.909  &.039  &.928  &.894  &.962  &.839  &.841  &.910  \\

\hline
\multirow{2}{*}{ECS.}
& \textbf{EL}&.905  &.062  &.909  &.890  &.948  &.868  &.846  &.917 \\
& \textbf{HL}&.836  &.044  &.934  &.918  &.972  &.909  &.899  &.955  \\
\hline
\multirow{2}{*}{HKU.}
& \textbf{EL}&.928  &.039  &.949  &.912  &.970  &.897  &.867  &.942 \\
& \textbf{HL}&.930  &.037  &.962  &.916  &.974  &.912  &.884  &.943  \\

\hline
\multirow{2}{*}{PAS.}
& \textbf{EL}&.877  &.079  &.855  &.844  &.917  &.796  &.788  &.879 \\
& \textbf{HL}&.894  &.065  &.894  &.879  &.943  &.846  &.845  &.902  \\

\hline
\multirow{2}{*}{SOD}
& \textbf{EL}&.786  &.124  &.811  &.763  &.881  &.801  &.757  &.849 \\
& \textbf{HL}&.806  &.112  &.836  &.779  &.912  &.825  &.777  &.878  \\
\hline
 \end{tabular}
 }
\label{table:DifferentLoss}
\end{table}

\begin{table}[t]
\caption{Salient object localization comparisons between our method and the conventional perceptual saliency models. The first three are our method with different schemes to assign $q$. CPD~\cite{wu2019cascaded}, MWS~\cite{zeng2019multi}, MLM~\cite{wu2019mutual}. ``P'': Precision, ``R'': Recall, ``F'': F-measure; ``DUT.'': DUTS-TE~\cite{zhao2015saliency}, ``ECS.'': ECSSD~\cite{yan2013hierarchical}, ``HKU.'': HKU-IS~\cite{li2016deep}, ``PAS.'': PASCAL-S~\cite{li2014secrets}, SOD~\cite{movahedi2010design}.}
\vspace{0.1cm}
\renewcommand\arraystretch{1.1}
\resizebox{1\linewidth}{!}{
\begin{tabular}{|c|rcccccc|}
\bottomrule

Datasets& &  $q$ Eq.~\ref{eq:q} & $q=3$ & $q=5$ & \textbf{CPD} & \textbf{MWS} & \textbf{MLM}\\
\hline
\hline
\multirow{3}{*}{DUT.}
& {Pre.} &.989 &.645 &.399 &.999 &.880 &.959 \\
& {Rec.} &.760 &.540 &.545 &.579 &.653 &.751 \\
& {Fme.} &.925 &.617 &.425 &.856 &.815 &.901 \\
\hline
\multirow{3}{*}{ECS.}
&{Pre.} &.843 &.520 &.470 &.706 &.755 &.769 \\
&{Rec.} &.964 &.786 &.949 &.851 &.660 &.916 \\
&{Fme.} &.868 &.564 &.532 &.735 &.730 &.798 \\
\hline
\multirow{3}{*}{HKU.}
&{Pre.} &.947 &.660 &.455 &.948 &.864 &.970 \\
&{Rec.} &.843 &.313 &.800 &.553 &.668 &.754 \\
&{Fme.} &.921 &.526 &.505 &.814 &.809 &.910 \\

\hline
\multirow{3}{*}{PAS.}
&{Pre.} &.888 &.788 &.694 &.871 &.652 &.884 \\
&{Rec.} &.949 &.913 &.931 &.783 &.907 &.891 \\
&{Fme.} &.902 &.814 &.738 &.849 &.697 &.885 \\

\hline
\multirow{3}{*}{SOD}
&{Pre.} &.916 &.446 &.303 &.906 &.795 &.747 \\
&{Rec.} &.942 &.738 &.813 &.890 &.875 &.864 \\
&{Fme.} &.922 &.490 &.355 &.902 &.812 &.771 \\
\hline
 \end{tabular}
 }
\label{table:DifferentAq}
\end{table}

\begin{table}[t]
\caption{Quantitative comparisons between our method and several classic fusion strategies. Max5, Mult5, and Mix5 respectively represent the saliency maps by fusing the 5 adopted saliency models using maximizing, multiplicative and average.}
\vspace{0.1cm}
\renewcommand\arraystretch{1.1}
 \resizebox{1\linewidth}{!}{
\begin{tabular}{|c|rcccccccc|}
\bottomrule
{\footnotesize Dataset}& & Sme.&  MAE &  ad.E& me.E& ma.E& ad.F& me.F& ma.F\\
\hline
\hline
\multirow{4}{*}{DUT.}
& \textbf{OUR}&.909&.039&.928&.894&.962&.839&.841&.910 \\
& \textbf{Mix5}&.873&.057&.848&.852&.926&.740&.783&.871 \\
& \textbf{Max5}&.820&.076&.782&.840&.899&.684&.721&.824 \\
&\textbf{Mult5}&.854&.056&.831&.879&.917&.732&.782&.843 \\
\hline
\multirow{4}{*}{ECS.}
& \textbf{OUR}&.836&.044&.934&.918&.972&.909&.899&.955\\
& \textbf{Mix5}&.926&.052&.920&.904&.961&.886&.879&.945\\
& \textbf{Max5}&.898&.054&.872&.916&.955&.859&.855&.928\\
&\textbf{Mult5}&.915&.046&.907&.929&.958&.888&.884&.932\\
\hline
\multirow{4}{*}{HKU.}
& \textbf{OUR}&.930&.037&.962&.916&.974&.912&.884&.943\\
& \textbf{Mix5}&.916&.045&.938&.900&.964&.878&.860&.931\\
& \textbf{Max5}&.886&.052&.904&.907&.947&.827&.831&.900\\
&\textbf{Mult5}&.899&.045&.921&.922&.957&.850&.856&.916\\

\hline
\multirow{4}{*}{PAS.}
& \textbf{OUR}&.894&.065&.894&.879&.943&.846&.845&.902\\
& \textbf{Mix5}&.856&.087&.834&.840&.905&.766&.790&.859\\
& \textbf{Max5}&.812&.100&.759&.834&.896&.749&.753&.844\\
&\textbf{Mult5}&.846&.082&.813&.865&.907&.784&.796&.861\\

\hline
\multirow{4}{*}{SOD}
& \textbf{OUR}&.806 &.112 &.836 &.779 &.912 &.825 &.777 &.878 \\
& \textbf{Mix5}&.797 &.120 &.815 &.766 &.878 &.792 &.758 &.848 \\
& \textbf{Max5}&.826 &.097 &.799 &.850 &.876 &.798 &.805 &.836 \\
&\textbf{Mult5}&.822 &.104 &.812 &.831 &.874 &.814 &.807 &.840 \\
\hline

 \end{tabular}
 }
\label{table:DifferentCombine}
\end{table}

\subsection{Component Evaluations}

\noindent$(A)$~\textbf{The effectiveness of using perceptual contrast (i.e., the 4096+4096 multi-scale deep feature in Sec.~\ref{sec:OSD})}. \\
We have tested the single scale deep feature performance in Table.~\ref{table:DifferentSubnet}, which indicates that our multi-scale perceptual contrast can effectively facilitate our semantic saliency re-ranking.
\vspace{0.2cm}

\noindent$(B)$~\textbf{The effectiveness of the hybrid image retrieval scheme (Sec.~\ref{sec:TDP})}. \\
In Table.~\ref{table:DifferentRetrieval}, we have tested the performance of using different image retrieval schemes, i.e., 1.Semantic Retrieval, 2.GIST+HOG, 3.Hybrid.
The results have suggested that the Hybrid retrieval scheme is more suitable to provide the stable object co-occurrence status for our semantic saliency learning.
\vspace{0.2cm}

\noindent$(C)$~\textbf{Why should we learn the semantic saliency relationship between objects rather than the semantic saliency for single object (Sec.~\ref{sec:NA})}?\\
We have respectively tested the performance of these two network choices in Table.~\ref{table:DifferentLoss}.
And the quantitative results have suggested to learn the semantic saliency relationship between objects, and the detailed discussion about this issue can be found in Sec.~\ref{sec:NA}.
\vspace{0.2cm}

\noindent$(D)~$\textbf{With regard to the salient object localization, can our learned semantic saliency outperform the conventional perceptual saliency (Sec.~\ref{sec:SSOL})}?\\
We have compared our method with the 3 most representative perceptual saliency models (i.e., CPD, MWS, MLM) toward the salient object localization in Table.~\ref{table:DifferentAq}, in which we have computed the precision, recall and f-measure based on the rectangular overlap rate between their saliency maps and GT.
The quantitative results have suggested that our method has achieved almost 8\% in f-measure improvement toward the salient object localization problem. Meanwhile, we have also verified the effectiveness of our adaptive scheme toward the adaptive choice of $q$ in Eq.~\ref{eq:q}.
\vspace{0.2cm}

\noindent$(E)$~\textbf{Since we have adopted 5 pre-trained saliency model in our saliency refinement component, can our method outperform the classic fusion strategies (i.e., directly fuse 5 saliency maps of these saliency models)}?\\
We have tried to fuse the saliency maps of the adopted 5 models via multiplicative, average, maximizing fusion strategies.
The quantitative data in Table.~\ref{table:DifferentCombine} have indicated that the simple combination of the 5 SOTA methods can not achieve an optimal complementary status between these methods.
And more qualitative comparisons can be found in Fig.~\ref{fig:ComponentDemo}.

\subsection{Limitation}
Our method tends to be time-consuming in general (Table.~\ref{table:Time}), because our approach is not an end-to-end model, and the object-level multi-scale deep feature computation is the major bottle-neck.

\begin{table}[t]
\caption{The detailed average time consumption (using a GTX1080GPU) toward different components in our method.}
\vspace{0.1cm}
\resizebox{1\linewidth}{!}{
\renewcommand\arraystretch{1.1}

\begin{tabular}{p{7.5cm} <{\centering}  p{2cm}<{\centering}}
\bottomrule

Main Steps& Time(seconds)\\
\hline
Object Proposal Preliminaries (Sec.~\ref{sec:OPP})&0.020\\
Multi-scale Deep Feature (Sec.~\ref{sec:OSD})&0.450\\
\hline
Semantically Salient Object Localization (Sec.~\ref{sec:SSOL})&0.150\\
Pixel-wise Saliency Refinement (Sec.~\ref{sec:PSR})&0.030\\
\hline
Total&0.650\\
\bottomrule
 \end{tabular}
}
\label{table:Time}
\end{table}
\section{Conclusion}
In this paper, we have revisited the problem of image salient object detection from the semantic perspective, i.e., the semantic saliency is more important than the conventional perceptual saliency to drive the attention of the real human visual system.
We have provided a novel network to learn the relative semantic saliency degree for any two objects.
Also, we have proposed a feasible weakly supervised scheme to train our siamese network for the learning of the semantic saliency.
Moreover, the final saliency map can be computed in the coarse-to-fine manner.
By ranking the inter-object semantic saliency degree, we have localized all those salient regions accurately, and then we have resorted 5 off-the-shelf deep models over these localized salient regions to achieve the pixel-wise saliency refinement.
We have conducted extensive quantitative evaluations to verify the effectiveness of each component in our method.

\vspace{0.2cm}
\noindent\textbf{Acknowledgments}. This research was supported in part by National Key R\&D Program of China (No. 2017YFF0106407), National Natural Science Foundation of China (No. 61802215 and No. 61806106), Natural Science Foundation of Shandong Province (No. ZR201807120086) and National Science Foundation of USA (No. IIS-1715985, IIS-0949467, IIS-1047715, and IIS-1049448).

\vspace{-0.2cm}
{\small
\bibliographystyle{unsrt}
\bibliography{CVPRMGX}
}

\end{document}